\documentclass{article}

\usepackage{graphicx}     
\usepackage{amsmath}
\usepackage{tikz}
\usepackage{interval}
\usepackage{amssymb}
\usepackage{placeins}

\newcommand{\W}{\mathcal{W}}
\newcommand{\U}{\mathcal{U}}

\newcommand\figref{Fig.~\ref}

\newcommand\tabref{Table~\ref}
\usepackage{booktabs}
\usepackage{amsmath}
\usepackage{authblk}
\usepackage{times}
\usepackage{geometry}
\usepackage{hyperref}

\title{Neural Network-Based Piecewise Survival Models}
\date{}
\author[1]{Olov Holmer}
\author[1]{Erik Frisk}
\author[1]{Mattias Krysander}

\affil[1]{Department of Electrical Engineering, Linköping University (email: name.lastname@liu.se)}

\begin{document}

\maketitle

\begin{abstract}
In this paper, a family of neural network-based survival models is presented. The models are specified based on piecewise definitions of the hazard function and the density function on a partitioning of the time; both constant and linear piecewise definitions are presented, resulting in a family of four models. The models can be seen as an extension of the commonly used discrete-time and piecewise exponential models and thereby add flexibility to this set of standard models. Using a simulated dataset the models are shown to perform well compared to the highly expressive, state-of-the-art energy-based model, while only requiring a fraction of the computation time.
\end{abstract}

\newcommand{\bin}[1]{{k\left(#1\right)}}
\newcommand{\bint}{\bin{t}}
\section{Introduction}

Survival analysis concerns the problem of describing the distribution of the time until an event occurs. It is used in a wide range of applications, for example in medicine \cite{flynn2012survival, katzman2018deepsurv} and predictive maintenance \cite{li2022attention, holmer2023energy, voronov2016heavy}. In many of these applications, it is of interest to predict the time of the event based on a set of covariates. In general, the relation between the covariates and the event time is complex and data-driven methods are therefore preferred; and like in many other fields, methods based on neural networks have shown to be particularly effective at this.

A big advantage of neural network-based survival models is that they can utilize the plethora of available neural network architectures that are constantly increasing, from simple feed-forward networks to specialized recurrent and transformer-based networks, and in that way handle a wide range of applications and corresponding covariates used as inputs. However, a survival model describes a statistical distribution and also needs to support censored data, which requires special consideration when designing the model. In \cite{wiegrebe2023deep}, 5 different classes of methods for doing this are identified, in which most available methods can be classified. The classes are: Cox-based, 
Discrete-time, 
parametric,
PEM-based, and
ODE-based.
The Cox-based approach is based on an extension of the Cox regression model where the log-risk function is modeled using a neural network \cite{faraggi1995neural}. Discrete-time methods assume a discrete time distribution and the discrete probabilities are based on the outputs from a neural network \cite{brown1997use,lee2018deephit}; this means that these models require the data to be from a discrete distribution, but they are often used on continuous data by binning the data based on a partitioning of the time \cite{kvamme2021continuous}. In parametric models, a neural network is used to parameterize a parametric distribution, often a Weibull distribution \cite{dhada2023weibull} or a mixture of Weibull distributions \cite{bennis2020estimation}. The PEM-based methods are based on a piecewise exponential model assumption where a neural network is used to parameterize the hazard rate \cite{kvamme2021continuous}. ODE-based methods utilize the fact that any nonnegative function can be used as a hazard function, the hazard function can therefore be directly parameterized using a neural network and the integration of the hazard function needed to calculate the survival function and hazard function can be seen as an ordinary differential equation for which there are methods available \cite{huang2020deepcompete,tang2022soden}; a more appropriate name for this class might be {\it integration-based methods} to also include the energy-based approach in \cite{holmer2023energy}.

The by far most well-used model classes are the Cox-based and the discrete-time methods. Cox-based methods rely on the proportional-hazards assumption which can be considered quite restrictive and often is not reasonable to assume, leaving the discrete-time method as the most well-used generally applicable method. A major drawback with the discrete-time methods is, however, the fact that they model a discrete distribution. In \cite{kvamme2021continuous} it is shown that by using interpolation, a discrete-time model can be converted to a continuous model; however, the results indicate that the PEM-model still yields better performance. The fact the PEM-model performs better is not surprising since it can be seen as an extension of the discrete-time methods where, instead of treating a continuous distribution based on binning the event times based on a partitioning of the time, a piecewise exponential model is defined on the partitioning. In this work, this idea is taken further by defining a family of models that are defined piecewise on a partitioning of the time.

\section{Preliminaries}
This section gives some preliminaries required to define the models.
\subsection{Survival Models}
Survival models are used to describe the distribution of a time $T$ until a specific event occurs. They are often described using the survival function 
\begin{equation}
   S(t\mid x) = P(T>t \mid x)
\end{equation}
where $x$ is the covariate vector, describing the probability that the event will occur after more than $t$ time units. In this work, we will only consider absolute continuous distributions, and in this case, the survival function can be specified using the density function $f$ as 
\begin{equation}
   S(t\mid x) = 1-\int_{0}^{t} f(s\mid x) \, ds.
\end{equation}
Alternatively, the survival function can be specified using the hazard rate
\begin{equation}
   h(t\mid x) = \lim_{\Delta t \rightarrow 0^+}\frac{P( t\leq T < t+\Delta t\mid T \geq t, x)}{\Delta t} =\frac{f(t\mid x)}{S(t\mid x)}
\end{equation}
using the cumulative hazard
\begin{equation}
   H(t\mid x) = \int_{0}^{t} h(s\mid x) \, ds
\end{equation}
as 
\begin{equation}
   S(t\mid x) = e^{-H(t\mid x)}.
\end{equation}

\subsection{Censoring}
When gathering survival data it is often not possible to record the event times of all individuals. This is because some individuals typically will drop out due to some unconsidered reason, or because the experiment ends before the event has happened. This means that the data contains right-censoring and the survival data from $N$ subjects have the form $\left\{(x_i,\tau_i,\delta_i) \right\}_{i=1}^N$ where, for each subject $i$, $\tau_i$ is the time of the event, and $\delta_i$ indicates if an event ($\delta_i=1$) or censoring ($\delta_i=0$) was observed.

\subsection{Maximum Likelihood Training}
The de facto standard for fitting survival models is maximum likelihood estimation. This is done by maximizing the likelihood, which fits well into the framework of training neural networks by simply using the log-likelihood as the loss function. 

Consider a model parameterized with $\theta$, the likelihood given a recorded failure is then
\begin{equation}
   L(\theta\mid \tau_i, x_i, \delta_i = 1) = p(T = \tau_i \mid \theta ,x_i) = f_\theta(\tau_i,x_i)
\end{equation}
and for a censored event it becomes
\begin{equation}
   L(\theta\mid \tau_i, x_i, \delta_i = 0) = P(T > \tau_i \mid \theta, x_i) = S_\theta(\tau_i, x_i).
\end{equation}
The total likelihood, for all $N$ individuals becomes
\begin{equation}\label{eq:likelihood}
   L\left(\theta\mid \{(x_i,\tau_i,\delta_i)\}_{i=1}^N\right) = \prod_{i\mid \delta_i=1} f_\theta(\tau_i, x_i) \prod_{i\mid \delta_i=0} S_\theta(\tau_i, x_i)
\end{equation}
where the products are taken over the sets of $i$ where $\delta_i=1$ and $\delta_i=0$, respectively. In practice, however, the logarithm of the likelihood is typically used. See \cite{holmer2023energy,kvamme2021continuous} for more details on this.

\section{Piecewise Survival Models}
In this section, four survival models that are defined piecewise using neural networks are presented, namely: the piecewise constant density model, the piecewise linear density model, the constant hazard model, and the linear hazard model. They are all defined using a discretization of the time between 0 and some maximal time $t_{max}$. The maximal time $t_{max}$ can be seen as a tuning parameter that should be chosen large enough so that all the recorded times in the dataset are smaller than $t_{max}$. 

When parameterizing a survival function it must be ensured that the resulting survival function is a proper survival function; that is, it must be a non-increasing function that starts at one ($S(0\mid x)=1$) and that goes to zero as the time goes to infinity ($\lim_{t\rightarrow \infty} S(t\mid x) =0$). Since we are only interested in the survival function on the interval $[0,t_{max}]$, the last condition, that the survival function goes to zeros, is replaced with the condition $S(t_{max}\mid x)>0$.

\subsection{Discretization Grid}
To define the piecewise models, a discretization of the time up to the maximal considered time $t_{max}$ is needed. Let  $\{\tau_i\}_{i=0}^N$ with $\tau_0=0 < \tau_1 < \cdots < \tau_N = t_{max}$ denote a grid with $N$ segments (and $N+1$ grid points). To simplify notation we define 
\begin{equation}
   k(t) = \max \left\{ i\in\left\{0,\ldots,N -1\right\} :  \tau_i  \leq t \right\}
\end{equation}
for $t_{min}\leq t \leq t_{max}$; i.e., the index of the segment that $t$ belongs to, so that $t\in \left[\tau_\bint, \tau_{\bint+1}\right)$. We also define 
\begin{equation}
   \Delta \tau_i=\tau_{i+1}-\tau_i
\end{equation}
for $i=1,\ldots N-1$. How to choose the grid points $\tau_i$ can be important for performance and in \cite{kvamme2021continuous} different ways of doing this are discussed. 

\subsection{Piecewise Constant Density Model}
In the piecewise constant density model, the density $f$ is a piecewise constant function with jumps at the grid times. The density is defined as 
\begin{equation}
   f(t\mid x) = f_\bint(x)
\end{equation}
where $f_i(x)$ denotes the value of the density function in segment $i$ (as a function of the covariate $x$), and the survival function becomes
\begin{equation}\label{eq:pcd_S}
   S(t\mid x) = 1-\int_{0}^{t} f(t)\, dt = 1- \sum_{i=0}^{k(t)-1}f_i \Delta\tau_i - (t-\tau_{k(t)}) f_{k(t)}.
\end{equation}
To parameterize this model using a neural network, let $z_0(x), \ldots, z_N(x)$ denote $N+1$ outputs from a neural network, evaluated at $x$, and let
\begin{equation}
   f_i(x) = \frac{e^{z_i(x)}}{e^{z_N(x)} + \sum_{j=0}^{N-1} \Delta\tau_j e^{z_j(x)}}.
\end{equation}
Using this parameterization,
\begin{equation}
   S(t_{max}\mid x) =  1- \sum_{i=0}^{N-1}f_i \Delta\tau_i = \frac{e^{z_N(x)}}{e^{z_N(x)} + \sum_{i=0}^{N-1} \Delta\tau_i e^{z_i(x)}}>0,
\end{equation}
which means that the last output, $z_N(x)$, from the neural network is used to determine the level of the survival function at $t_{max}$. By design, we also have $S(0)=1$ and $f(t\mid x)>0$ and consequently $S$ is a proper survival function.

\subsection{Piecewise Linear Density Model}
In this model, the density is modeled as a continuous piecewise linear function as
\begin{equation}
   f(t\mid x) = f_\bint(x) + \frac{t-\tau_\bint}{\Delta \tau_\bint}\left(f_{\bint+1}(x)-f_\bint(x)\right)   
\end{equation}
where $f_i(x)$ denotes the value of the density at the grid time $\tau_i$.  Using this density, for any $t \in [\tau_i, \tau_{i+1}]$
\begin{equation}
   \begin{split}
   \int_{\tau_{i}}^{t} f(t\mid x)\,dt &= \int_{\tau_{i}}^{t} \left(f_i(x) + \frac{t-\tau_i}{\Delta \tau_i}\left(f_{i+1}(x)-f_i(x)\right)\right) \,ds 
   =
   \left[ f_i(x) s + \frac{\left(s-\tau_i\right)^2}{2\Delta \tau_i}\left(f_{i+1}(x)-f_i(x)\right) \right]_{\tau_i}^t
   \\
   &= f_i(x)(t-\tau_i) + \frac{\left(t-\tau_i\right)^2}{2\Delta \tau_i}\left(f_{i+1}(x)-f_i(x)\right)
   \end{split}
\end{equation} 
and in particular
\begin{equation}
   \begin{split}
   \int_{\tau_{i}}^{\tau_{i+1}} f(t\mid x)\,dt = \frac{\Delta \tau_i}{2}\left(f_{i+1}(x)+f_i(x)\right).
   \end{split}
\end{equation} 
The survival function can therefore be written 
\begin{equation}\label{eq:intergral_linear_f}
   \begin{split}
      S(t\mid x) &= 1-\int_{0}^{t} f(t)\, dt \\&= 1- \left(\sum_{i=0}^{k(t)-1} \frac{\Delta \tau_i}{2}\left(f_{i+1}(x)-f_i(x)\right)\right) - 
      f_{k(t)}(x)(t-\tau_{k(t)}) - \frac{\left(t-\tau_{k(t)}\right)^2}{2\Delta \tau_{k(t)}}\left(f_{k(t)+1}(x)-f_{k(t)}(x)\right).
   \end{split}
\end{equation}
Similarly as for the constant density model, a neural network with $N+2$ outputs $z_0(x), \ldots, z_{N+1}(x)$ can be used to parameterize the density as 
\begin{equation}
   f_i(x) = \frac{e^{z_i(x)}}{e^{z_i(x)} +\sum_{j=0}^{N-1} \frac{\Delta \tau_i}{2}\left(e^{z_{i+1}(x)}+e^{z_i(x)}\right)}. 
\end{equation}
Using this density we have 
\begin{equation}
   S(t_{max}\mid x)= 1- \sum_{i=0}^{N-1} \frac{\Delta \tau_i}{2}\left(f_{i+1}(x)-f_i(x)\right) = \frac{e^{z_{N+1}(x)}}{e^{z_i(x)} +\sum_{j=0}^{N-1} \frac{\Delta \tau_i}{2}\left(e^{z_{i+1}(x)}+e^{z_i(x)}\right)},
\end{equation}
and again we have a proper survival function where the last output $z_{N+1}(x)$ from the neural network determines the level of the survival function at time $t_{max}$.
\subsection{Piecewise Constant Hazard Model}
In this model, the hazard rate is modeled as a piecewise constant function as 
\begin{equation}
   h(t\mid x) = h_{k(t)}(x)
\end{equation}
where $h_i(x)$ is the hazard rate in segment $i$; and since any nonnegative function is a proper hazard rate, this model can directly be parameterized by a neural network with $N$ outputs $z_0,\ldots,z_{N-1}$ as 
\begin{equation}
   h_i(x) = e^{z_i(x)}.
\end{equation}
Note that no extra output from the network is needed to represent the value of the survival function at time $t_{max}$ using this parameterization.

The cumulative hazard for this model is
\begin{equation}
   H(t\mid x) = \int_{0}^{t} h(t\mid x) = \sum_{i=0}^{k(t)-1} \Delta t_i h_i(x) + (t-\tau_{k(t)})h_{k(t)}(x),
\end{equation}
based on which the survival function can be calculated as $S(t\mid x) = e^{-H(t\mid x)}$. Finally, the density can be calculated as $f(t\mid x) = S(t\mid x) h(t\mid x)$.

\subsection{Piecewise Linear Hazard Model}
Similarly to the linear density model, this model uses a continuous piecewise linear function to parameterize the hazard rate. The hazard rate is defined as 
\begin{equation}
   h(t\mid x) = h_\bint(x) + \frac{t-\tau_\bint}{\Delta \tau_\bint}\left(h_{\bint+1}(x)-h_\bint(x)\right)   
\end{equation}
where $h_i(x)$ denotes the value of the hazard rate at the grid time $\tau_i$. As in the piecewise constant model, the hazard can be directly parameterized using a neural network, but in this case with $N+1$ outputs $z_0,\ldots,z_{N-1}$, as 
\begin{equation}
   h_i(x) = e^{z_i(x)}.
\end{equation}
Based on \eqref{eq:intergral_linear_f} the cumulative hazard can be written
\begin{equation}
   \begin{split}
      H(t\mid x) &= \int_{0}^{t} h(t\mid x)\, dt 
      \\ &=
      \left(\sum_{i=0}^{k(t)-1} \frac{\Delta \tau_i}{2}\left(h_{i+1}(x)-h_i(x)\right)\right) - 
      h_{k(t)}(x)(t-\tau_{k(t)}) - \frac{\left(t-\tau_{k(t)}\right)^2}{2\Delta \tau_{k(t)}}\left(h_{k(t)+1}(x)-h_{k(t)}(x)\right)
   \end{split}
\end{equation}
and the survival function $S$ and the density $f$ can be calculated in the same way as for the piecewise constant hazard model.

\subsection{Some Notes on the Implementation of the Models}
The models above are all written in a form that should be straightforward to implement in modern neural network software. However, there are a few details worth mentioning, especially when it comes to evaluating the log-likelihood. In particular, by applying the following relations
\begin{align}
   \log z_1 z_2 &= \log z_1 +\log z_2\\
   \log e^{z} &= z\\
   \log \left(\sum_{i} e^{z_i}\right) &=  \log \left(e^{\max_i z_i} \sum_{i} e^{z_i-\max_i z_i}\right) =  \max_i z_i + \log \left(\sum_{i} e^{z_i-\max_i z_i}\right) 
\end{align}
the equations can be simplified quite significantly, and also better numerical stability is achieved. These simplifications are however quite straightforward and are therefore omitted here.  

\section{Comparison of the Models}
In this section, the four presented models are compared with each other as well as with the energy-based model presented in \cite{holmer2023energy}, using the simulated dataset in \cite{holmer2023energy}. The simulated data is drawn from a two-parameter Weibull distribution with the survival function
\begin{equation}
   S_{\W(\lambda,k)}(t) = e^{-\left( \frac{t}{\lambda} \right)^{k}}
\end{equation}
where $k > 0$ is the shape parameter and $\lambda > 0$ is the time-scale parameter of the distribution. To create the dataset of $N$ individuals, for each individual $i$ the parameters are drawn from uniform distributions, according to $\lambda_i \sim \U(1,3)$ and $k_i \sim \U(0.5,5)$, and the covariate vector is taken as $x_i=[\lambda_i, k_i]$. Three different datasets were used, one for training, one for validation and one for testing the model after training, of sizes 1000, 300, and 300, respectively.

A fully connected feed-forward network was used in all models, and it was found that two layers of 32 nodes each and a uniformly spaced grid with 5 grid points was sufficient for all models, in the sense that increasing the number of nodes or grid points did not yield better results. 20 different learning rates geometrically spaced between $10^{-1}$ and $10^{-4}$ were evaluated for each model and the best performing one was selected. Each model was trained for 200 epochs and the parameters from the epoch with the lowest loss on the validation set were taken as the resulting model from the training; no other regularization was used; however, dropout and weight decay were evaluated but not used in the end since they did not yield better performance. The hyperparameters and training of the energy-based model were the same as in \cite{holmer2023energy}.

The results from the comparison are summarized in \tabref{table:results}. As can be seen, the piecewise linear models perform better than the piecewise constant ones, which is as expected since the true survival function is a smooth function. It can also be seen that the hazard-based parameterizations perform better than the density-based parameterizations, a possible reason for this is that the hazard of the true distribution is a monotonically increasing function while the density function has a bell shape. Slightly simplified one could say that varying the covariate vector mainly affects the slope of the hazard while both the with and location of the peak in the density will change, which makes it easier to parameterize the hazard using a neural network. It can also be seen that the performances of the best-performing piecewise models are similar to that of the energy-based model, but they only require a fraction of the training time of the energy-based model.  

\begin{table}[h!]
   \centering
   \begin{tabular}{l | r r} 
    \hline
    Model & Test loss & Training time [s] \\ [0.5ex] 
    \hline\hline
    Constant Density  &  0.607 $\pm$  0.0699 & 0.603 $\pm$ 0.1068 \\ 
    Constant Hazard  &   0.6 $\pm$  0.07138& 0.596 $\pm$ 0.08736 \\ 
    Linear Density  & 0.582 $\pm$  0.0644 & 0.691 $\pm$ 0.08302 \\ 
    Linear Hazard  & 0.561 $\pm$  0.0592 &0.671 $\pm$ 0.08249 \\ 
    Energy Based  & 0.549 $\pm$  0.0615 & 2.16 $\pm$ 0.2243\\ [1ex] 
    \hline
   \end{tabular}
   \caption{Results from training the models on 100 different representations of the simulated dataset. The values before and after the $\pm$ represent the mean and standard deviation, respectively.}
   \label{table:results}
   \end{table}

\begin{figure}[h]
      \centering
      \includegraphics[]{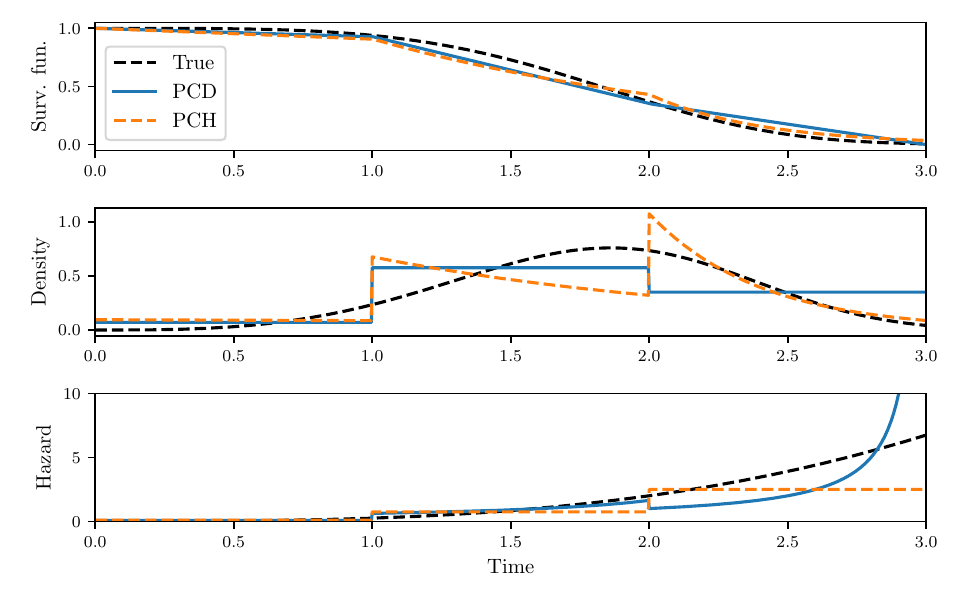}
      \caption{Comparison of the piecewise constant models for $\lambda=2$ and $k=3$. The number of grid points is only 3 to make the differences in appearance of the models more clear.}
      \label{fig:1}
  \end{figure}

\begin{figure}[h]
   \centering
   \includegraphics[]{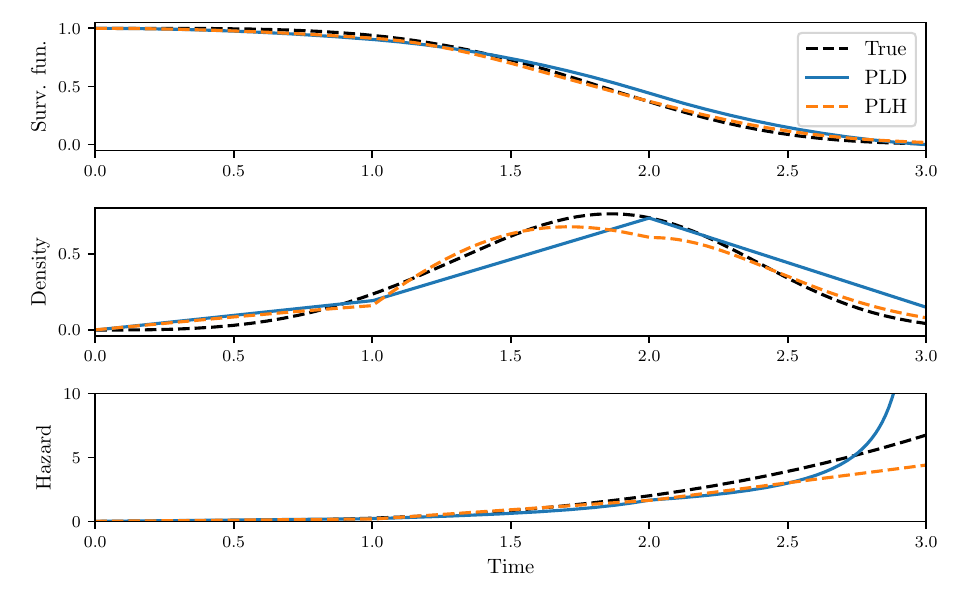}
   \caption{Comparison of the piecewise linear models for $\lambda=2$ and $k=3$. The number of grid points is only 3 to make the differences in appearance of the models more clear.}
   \label{fig:2}
\end{figure}

In \figref{fig:1} and \figref{fig:2} the density formulations are compared with the hazard formulations for a specific covariate vector. In these figures, it can be seen that the density-based parameterization gives a piecewise linear and piecewise quadratic survival function, respectively, which results in a predictable behavior that is desirable in many cases, while the hazard function is less predictable. For example, the constant-hazard model gives a piecewise exponential survival where the slope of the survival function is higher at the beginning of each interval corresponding. On the other hand, while the hazard-based parameterization can be considered to have an appearance that is not as easy to interpret, the linear hazard formulation gives a survival function and density function that is very close to the ground truth. 

\section{Conclusion}
In this paper, a family of neural network-based survival models is presented. The models are specified based on piecewise definitions of the hazard function and the density function on a partitioning of the time. Both constant and linear piecewise definitions are presented resulting in a total of four models: piecewise constant density, piecewise linear density, piecewise constant hazard, and piecewise linear hazard.

The models are compared using a simulated dataset and the results show that the piecewise linear models give better performance than the piecewise constant ones. The results also indicate that models based on the parameterization of the hazard function give better performance; however, this likely depends on the considered dataset. The models are also compared to the highly expressive energy-based model and the piecewise linear models are shown to have similar performance to this model but only require a fraction of the calculation time.

\bibliographystyle{plain} 
\bibliography{refs}

\end{document}